%% file: main.tex
\documentclass[letterpaper, 10 pt, conference]{ieeeconf}  

\IEEEoverridecommandlockouts                              

\usepackage{amsmath}
\usepackage{amssymb}
\usepackage{float}
\usepackage{graphicx}
\usepackage{tabularx}
\usepackage{booktabs}
\usepackage{cite}
\usepackage{url}
\usepackage{caption}
\usepackage{subcaption}
\usepackage{flushend} 
\usepackage{hyperref}

\input{macros}

\title{\LARGE \bf
Offline Adaptation of Quadruped Locomotion using Diffusion Models
}
\author{Reece O'Mahoney, Alexander L. Mitchell, Wanming Yu, Ingmar Posner, Ioannis Havoutis
\thanks{Authors are with the Oxford Robotics Institute, Department of Engineering Science,
        University of Oxford,
        {\tt\small \{reeceo, mitch, wanming, ingmar, ioannis\}@robots.ox.ac.uk}}%
        }

\begin{document}
\maketitle
\thispagestyle{empty}
\pagestyle{empty}
\begin{abstract}

We present a diffusion-based approach to quadrupedal locomotion that simultaneously addresses the limitations of learning and interpolating between \emph{multiple} skills (modes) and of \emph{offline adapting} to new locomotion behaviours after training. This is the first framework to apply classifier-free guided diffusion to quadruped locomotion and demonstrate its efficacy by extracting goal-conditioned behaviour from an originally unlabelled dataset. We show that these capabilities are compatible with a multi-skill policy and can be applied with little modification and minimal  compute overhead, i.e., running entirely on the robot’s onboard CPU. We verify the validity of our approach with hardware experiments on the ANYmal quadruped platform. Code and videos are available at \url{https://reeceomahoney.vercel.app/publications/locodiff}.
\end{abstract}

\section{Introduction}

Quadruped robots' ability to traverse complex terrain while carrying useful payloads makes them excellent choices for applications in manufacturing, construction and search \& rescue. The state of the art for quadruped locomotion is maturing rapidly and both learning-based and gradient based methods are prevalent in research and industry. Learning-based methods such as reinforcement learning (RL) have produced impressive results~\cite{hwangbo2019learning,gangapurwala2020guided,gangapurwala2020rloc,rudin2022learning,hoeller2023anymal,caluwaerts2023barkour}, but still suffer from a number of limitations. Due to the limited expressive capacity of neural network policies, using multiple skills in previous works required a hierarchical approach with a difficult multi-step training pipeline~\cite{hoeller2023anymal}. Since the methods are trained online, the resultant policies often can't be adapted to new behaviours without being retrained from scratch. Our approach attempts to propose a solution to each of these two limitations in particular.

To alleviate these issues, an alternative learning-based paradigm, imitation learning, can be used. This involves reproducing a set of reference motions, which popular approaches achieve with an adversarial training loop \cite{peng2021amp}. 
In recent years, Diffusion models~\cite{ho2020denoising,huang2024diffuseloco} have outperformed prior methods such as \cite{gail} and are able to approximate multi-modal distributions at a high fidelity. By approximating the score function \cite{hyvarinen2005} of a distribution, instead of directly learning the densities, and iteratively applying Langevin dynamics \cite{song2020generative} to generate samples, they are able to accurately represent highly multi-modal distributions, incorporate flexible conditioning \cite{janner2022planning}, and also scale well with dataset and model size. This multi-modality allows us to circumvent the need for hierarchical planning, as previous works have shown diffusion policies are capable of learning multiple skills in a single model \cite{huang2024diffuseloco}.

\begin{figure}
    \centering
    \includegraphics[width=\linewidth]{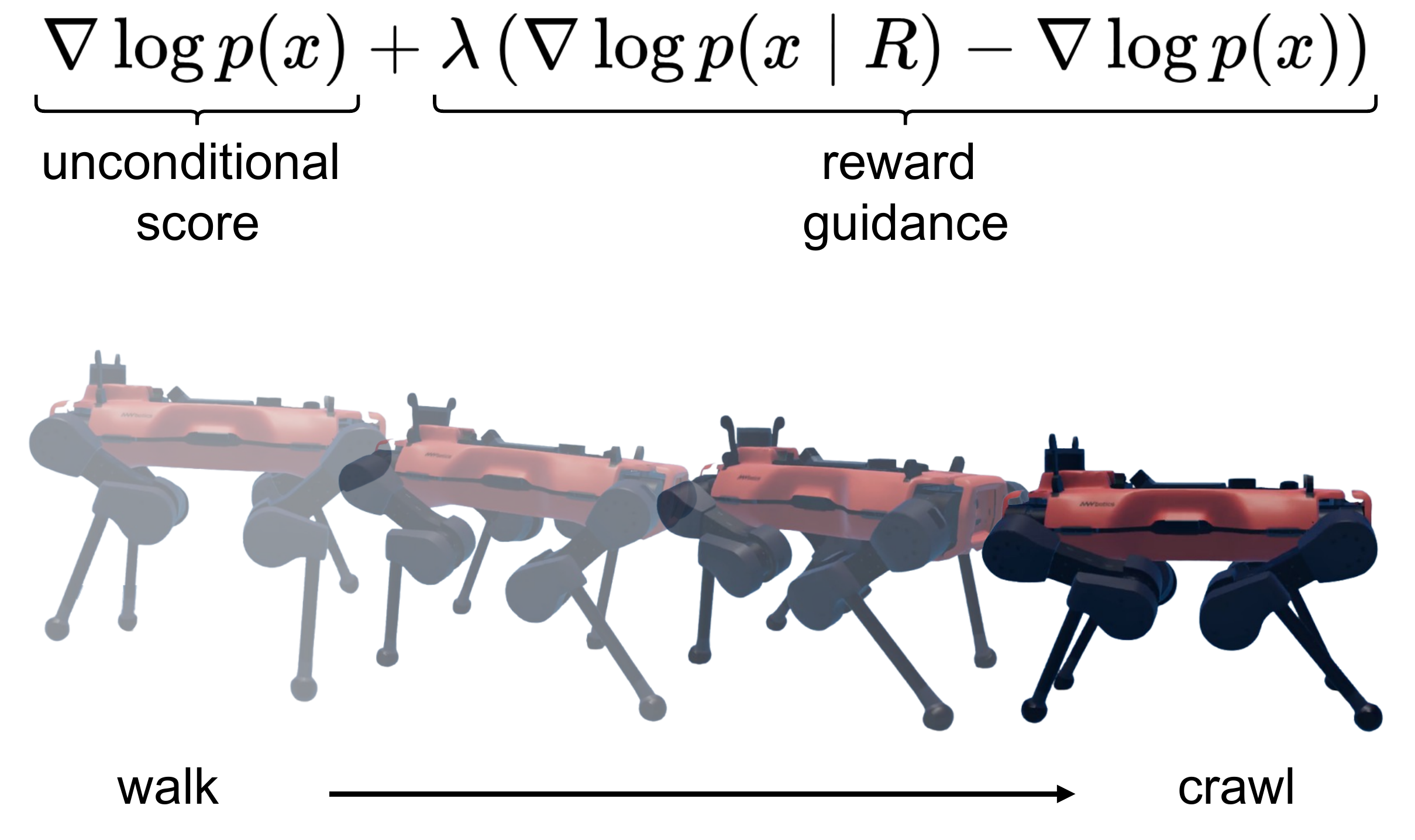}
    \caption{Our diffusion-based approach possesses two capabilities: interpolation between distinct skills or operation modes (walk to a low crawl) and offline adaption after training. We leverage diffusion models to approximate multi-modal distributions to a high fidelity and also show that using classifier-free guidance can achieve offline adaption after training to produce novel locomotion behaviours. The guidance is used to generate trajectories that maximise the return of new reward functions unseen during data collection.}
    \vspace{-0.65cm}
    \label{fig:abstract}
\end{figure}

Though large datasets are prevalent and can fuel imitation learning, there are limitations.
There is no guarantee that the dataset contains trajectories which maximise a desired test time reward. For example, our locomotion dataset may contain a broad range of motions where the robot is moving with different gaits and speeds, but at test time we require specific behaviours. If data is labelled it is easy to extract this desired behaviour, but this is often not the case making this problem challenging. Therefore, we would like to tune our diffusion model to stitch together locomotion trajectories which maximise the return of reward functions for certain types of motion. 

To tackle the challenge of offline adaptation, we propose using classifier-free guidance (CFG)~\cite{ho2022classifierfree} to optimise a diffusion-model trajectories after training. This is achieved by treating the expected return as a probability function \cite{janner2022planning, ajay2023conditional} and using an Offline RL~\cite{levine2020offlinereinforcementlearningtutorial} training paradigm. 
In particular, we label the dataset with a velocity tracking reward and use CFG to guide our trajectories to maximise it, thus adapting our model to generate the desired behaviour.

In this paper we present a diffusion-based approach to quadruped locomotion, capable of switching between discrete skills and adapting to new reward functions offline. 

Our contributions are as follows:
\begin{enumerate}
    \item We present the first application of a stochastic differential equation-based diffusion framework to quadruped locomotion and use it to train a policy capable of interpolating between discrete skills.
    \item We apply classifier-free guidance to perform Offline RL and adapt our policy to new goal-conditioned behaviour.
    \item We deploy onto real hardware and demonstrate the benefits of fast sampling with a model that can run entirely on the robot's onboard CPU.
\end{enumerate}

We evaluate our approach on a set of locomotion tasks both in simulation and on hardware using an ANYbotics ANYmal quadruped robot~\cite{ANYmal}.

\section{Related work}

\textbf{\textit{Learning-based approaches to locomotion}} - Advancements in reinforcement learning have demonstrated a remarkable ability for learning highly dynamic locomotion policies \cite{caluwaerts2023barkour,hoeller2023anymal}. When handling multiple skills, most state of the art methods have opted to for a hierarchical approach, where individual skills are learnt with different policies, and a high level policy is trained to switch between them based on the agents state and terrain features \cite{hoeller2023anymal}. Despite the high-performance ceiling of this type of method, this formulation limits the expressive capability of these models to learn motions between these skills or outside of them. Being able to handle multiple skills in a single model would be preferable both for simpler training and also to allow continuous interpolations between them as opposed to discrete switching~\cite{mitchell2024gaitor}.

Another approach to learning locomotion skills is imitation learning, where instead of designing some reward function, an agent is trained to mimic some reference clip, typically generated from motion capture data. One popular approach in this category is adversarial motion priors (AMP) \cite{peng2021amp}. While AMP has been shown to be capable of handling multiple skills in a quadruped locomotion setting \cite{vollenweider2022multipleadversarial}, it still has a few disadvantages, namely that it is adversarial method which notoriously suffer from scalability problems due to training instability \cite{gail}. Our work could also be seen in the context of policy distillation, which aims to combine multiple policies into one without the need for hierarchical methods. One popular approach to this is DAgger~\cite{ross11a2011dagger} which works by collecting a dataset using current policy at each iteration and updating it using the whole dataset.

\textbf{\textit{Diffusion for control}} - Due to their ability to accurately model highly multi-modal distributions, diffusion models have become an extremely popular paradigm for generating high fidelity images and video~\cite{rombach2022highresolution} as well as control trajectories~\cite{ajay2023conditional, chi2024diffusion, janner2022planning}. These models exhibit several appealing properties including high accuracy, allowing for easy deployment of policies trained offline~\cite{chi2024diffusion}, an ability to generate variable length trajectories in a single step, allowing for long-horizon predictions that do not suffer from compounding error~\cite{janner2022planning}, and an amenability to flexible conditioning by either fixing points in the trajectory~\cite{janner2022planning}, guiding generation using reward functions, or enforcing goals and/or constraints~\cite{ajay2023conditional, reuss2023goal}. While one other prior work has looked at performing Offline RL using CFG, our work differs in the application of this method beyond simple toy problems and attempts to scale it up to real-world dynamics.

To date, most work has either only applied this framework to simulation environments and simple maze problems~\cite{janner2022planning}. Deployment on hardware has been mostly limited to manipulation tasks~\cite{chi2024diffusion}, with only one other work~\cite{huang2024diffuseloco} deploying to a quadruped. Our work however has a number of key differences. Instead of purely focusing on the multi-skill capabilities of diffusion models we perform offline RL by applying return conditioned classifier-free guidance. As well as this, they use the Denoising Diffusion Probabilistic Models (DDPM)~\cite{ho2020denoising} framework, as opposed to ours which leverages multiple improvements~\cite{song2021score, karras2022elucidating, lu2023dpmsolver} to the generative process by casting it in the framework of stochastic differential equations (SDEs), allowing for the best samplers to be chosen at inference time without retraining, and crucially for robotics, faster inference through deterministic sampling, requiring only diffusion 3 steps as opposed to their 10, to generate high quality motions. This enabled us to run our policy using only the onboard CPU instead of requiring specific hardware acceleration.

\textbf{\textit{Offline reinforcement learning}} - As opposed to online learning which requires access to a high fidelity simulator, offline learning aims to extract behaviour directly from pre-collected datasets, often of mixed quality~\cite{levine2020offlinereinforcementlearningtutorial}. There have been several popular strategies to date including constraining value estimations outside of distribution~\cite{kumer2020conversative}, rewarding the policy for staying close to the behaviour policy's state distribution~\cite{fujimoto2021minimalist, wang2023diffusion}, and implicitly learning a value function via inter-quartile regression~\cite{kostrikov2022offline, hansen-estruch2023implicit}. There has been little work applying Offline RL to quadruped locomotion but recent work has proposed a benchmark for this area in an effort to encourage more research~\cite{zhang2023realworldquadrupedallocomotionbenchmark}.

\section{Method}

\begin{figure*}
    \centering
    \begin{subfigure}[b]{0.3\textwidth}
        \centering
        \includegraphics[height=5cm]{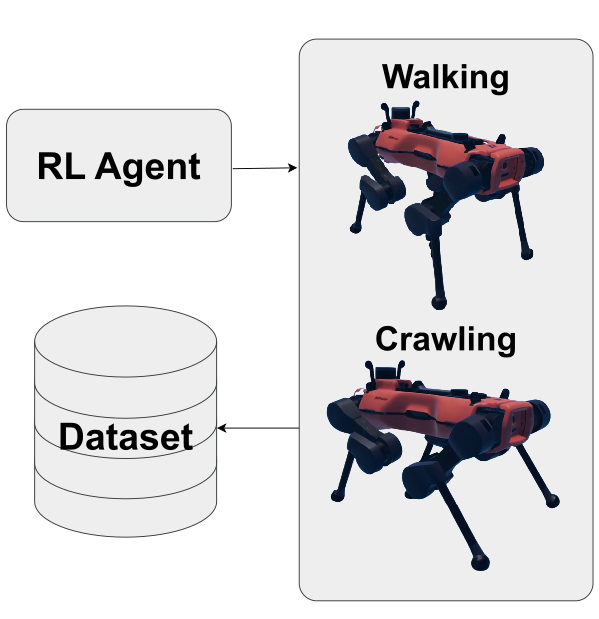}
        \caption{Dataset collection}
    \end{subfigure}
    \begin{subfigure}[b]{0.4\textwidth}
        \centering
        \includegraphics[height=5cm]{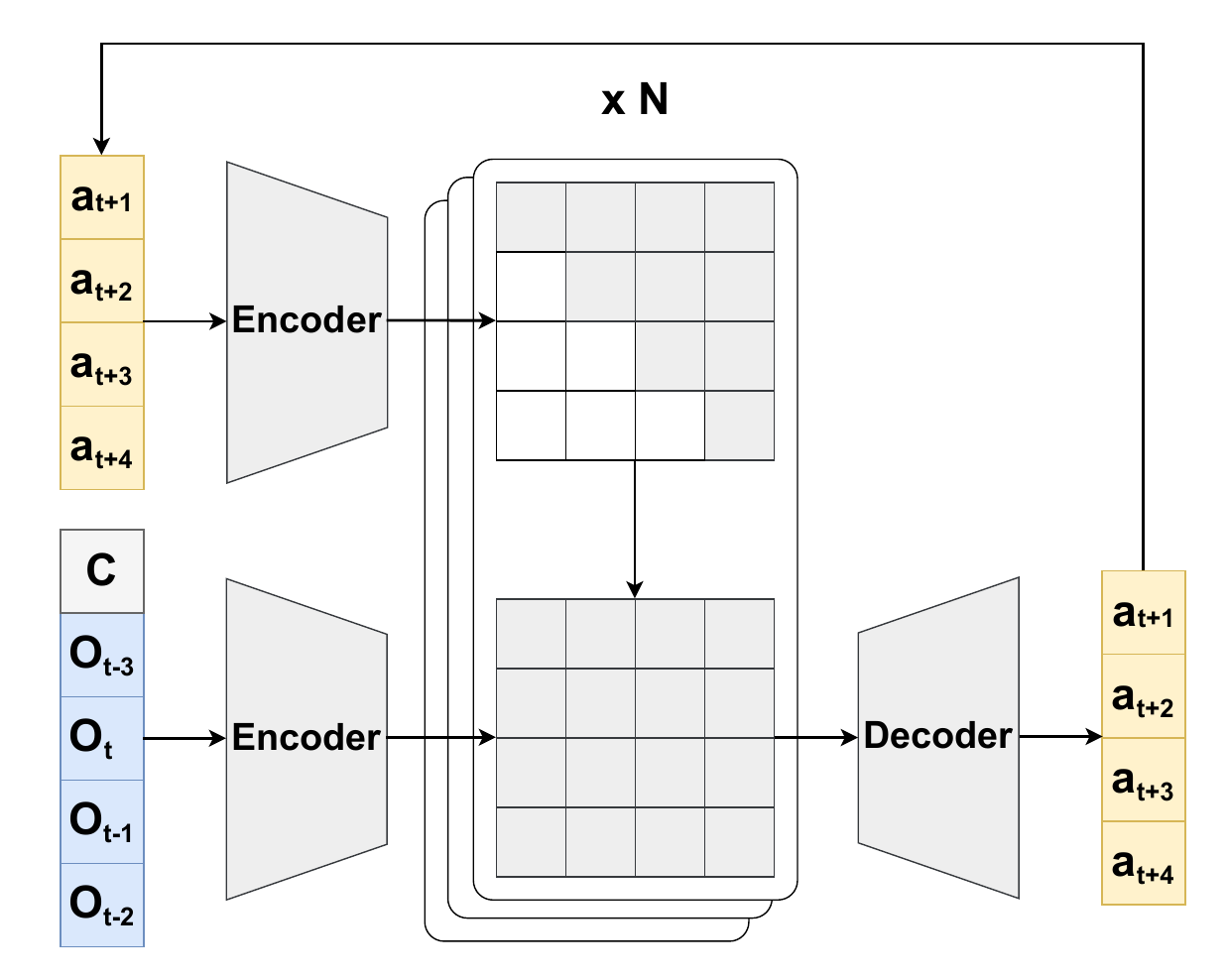}
        \caption{Training}
    \end{subfigure}
    \begin{subfigure}[b]{0.28\textwidth}
        \centering
        \includegraphics[height=5cm]{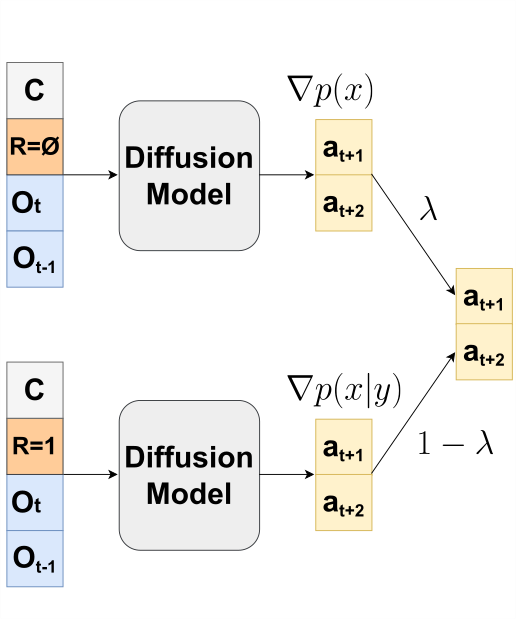}
        \caption{Reward guidance}
    \end{subfigure}
    \caption{Method Overview: a) A reinforcement learning agent is pre-trained with a hand crafted policy that generates reference trajectories. These are collected by rolling out the policy in an environment with randomised parameters for robustness. b) An embedding of the observation is concatenated with separate embeddings of the diffusion timestep, skill, and return. These together form the conditioning input. The multi-head transformer decoder initially takes a noise vector as input, applies causal self-attention, then cross-attention with the conditioning and produces a partially denoised vector. This process is repeated N times to produce a complete action trajectory. c) The return value is randomly masked during training, allowing us to use classifier-free guidance at test time. This is done by taking a weighted sum of unconditional and maximum return trajectories at each denoising step.}
    \label{fig:overview}
    \vspace{-0.5cm}
\end{figure*}

Our method trains a diffusion model over a diverse range of locomotion behaviours. A classifier-free guidance (CFG) technique for adapting the output of the diffusion model to optimise for specific test-time rewards is also presented. Lastly, we explain the specifics of our architecture that were required to apply this model to locomotion.

\subsection{Diffusion formulation}
Diffusion models were independently derived by two different lines of research~\cite{ho2020denoising,song2020generative} with two corresponding interpretations. In this work we focus on the score-based formulation introduced in~\cite{song2020generative}. The score function of a distribution is the gradient of its log density function. This can be learnt by several approaches~\cite{vincent2010score, song2019sliced} but most works use the denoising score matching loss below, where \(\bs_\theta(\tilde{\xx})\) is a neural network, \(p(\xx)\) is the data distribution and \(p(\tilde{\xx})\) is the noise perturbed distribution:
\begin{equation}
 \frac{1}{2}\mathbb{E}_{q_\sigma(\tilde{\xx}\mid \xx)p_\text{data}(\xx)}[\| \bs_\theta(\tilde{\xx}) -                       \nabla_{\tilde{\xx}} \log q_\sigma(\tilde{\xx}\mid \xx) \|_2^2]. \label{eqn:dsm}
\end{equation}
The authors in~\cite{song2020generative} make the additional leap that by learning this score function for multiple noise perturbation scales between our data distribution \(p(\xx_0)\) and some prior distribution \(p(\xx_T)\), we can generate samples from \(p(\xx)\) by using Langevin dynamics to iteratively step in the reverse direction. This uses the following equation:
 \begin{equation}
 \tilde{\xx}_{t} = \tilde{\xx}_{t-1} + \frac{\epsilon}{2}\nabla_{\xx} \log p(\tilde{\xx}_{t-1}) +  \sqrt{\epsilon}~ \zz_t,   \label{eqn:langevin}
\end{equation}
where \(\zz_t\) is Gaussian noise. It was shown in~\cite{song2021score} that this discrete process can be generalised into a continuous one by using the Ito SDE. Doing this allows for greater performance, decouples the number of inference steps from the training procedure and, most importantly for real-time control, allows for fast sampling. The latter is due to the fact that a probability flow ODE can be derived which has the same marginal distributions as the original SDE. Since this process has no additive noise \(\zz_t\) during sampling, much fewer inference steps are needed as the effect of this noise does not need to be corrected for. In this work we leverage the k-diffusion framework~\cite{karras2022elucidating}, which builds upon the above. In particular, we define our ODE using the following equation:
\begin{equation}
\diff \xx = -\dot\sigma(t) ~\sigma(t) ~\nablaxx \log p \big( \xx; \sigma(t) \big) ~\diff t \text{,}
\end{equation}
where \(\sigma(t)\) defines the variance schedule of our perturbation kernel as a function of \(t\in[0,1]\). The  score function is not predicted directly, but instead through the following preconditioning:
\begin{equation}
\label{eq:scaling}
    D_{\theta}(\xx, \sigma) = c_{skip}(\sigma)\xx + c_{out}(\sigma)F_{\theta}(c_{in}(\sigma)\xx ; c_{noise}(\sigma)),
\end{equation}
where \(F_{\theta}\) is our network output, \(D_{\theta}\) is the final denoising output, and the remaining parameters are scaling factors applied to the input and output, as calculated in~\cite{karras2022elucidating} to stabilise training and minimise prediction error. This has the benefit of allowing our network to more easily learn at multiple noise levels, as it can either predict the score function directly when \(\sigma\) is large or a residual term that is subtracted from the current \(\xx\) when \(\sigma\) is small.

\renewcommand{\arraystretch}{1.5}
\begin{table*}[t]
    \centering
    \begin{tabularx}{\linewidth}{X*{4}{>{\centering\arraybackslash}X}}
        \toprule
         Model & \(v_x = 0.8\) & \(v_x = -0.8\) & \(v_y = 0.5\) & \(\omega_z = 1.0\) \\
         \midrule 
         Expert model & 0.87 $\pm$ 0.09 & 0.89 $\pm$ 0.08 & 0.90 $\pm$ 0.07 & 0.87 $\pm$ 0.09 \\
         Ours (DDPM) & 0.84 $\pm$ 0.11 & 0.84 $\pm$ 0.12 & 0.86 $\pm$ 0.09 & 0.85 $\pm$ 0.10 \\
         Ours (DDIM) & 0.84 $\pm$ 0.11 & 0.85 $\pm$ 0.12 & 0.85 $\pm$ 0.10 & 0.86 $\pm$ 0.09 \\
         Ours (DPM++(2M)) & 0.74 $\pm$ 0.15 & 0.73 $\pm$ 0.15 & 0.77 $\pm$ 0.13 & 0.72 $\pm$ 0.15 \\
         \bottomrule
    \end{tabularx}
    \caption{Average velocity tracking reward for different models and reward functions. Expert model has access to the true velocity command and is conditioned using these during training. The last three are our same model, trained with the SDE framework, but with different samplers chosen at inference time. Results are averaged over 100 environments with 250 steps each. Results are taken at the value of lambda that gave the best rewards.}
    \label{tab:model_comp}
    \vspace{-0.5cm}
\end{table*}
\renewcommand{\arraystretch}{1}

We train the denoiser by predicting the original value of noise-corrupted samples from the dataset. The noise levels are drawn from a log-logistic distribution as this has been shown to produce better performance. The loss for a given \(\sigma\) thus takes the following form:
\begin{multline}
\mathbb{E}_{\yy \sim p_{data}} \mathbb{E}_{\nn \sim \mathcal{N}(\boldsymbol{0}, \sigma^2 \boldsymbol{I})} 
\lVert D(\yy + \nn;\sigma) - \yy \rVert^2_2 
\text{,} \\ \hspace*{2mm}\text{where}\hspace*{2mm}
\nablaxx \log p(\xx; \sigma) = \big( D(\xx; \sigma) - \xx \big) / \sigma^2.
\end{multline}

During sampling we test a number of different schemes that were shown to produce the best performance on different tasks in prior work~\cite{reuss2023goal}. We chose to use 3 denoising steps at inference time as we found the extra performance from additional steps was marginal compared to the increase in inference time.

\subsection{Classifier-free guidance}
\label{sec:cfg}

By applying Bayes rule and the definition of the score function to the conditional distribution \(p(x \mid y)\), we can produce the following equation for Classifier-Free Guidance:

\begin{multline}
\label{eq:cfg}
    \nabla_x \log p(x \mid y) = \\ (1 - \lambda) \nabla_x \log p(x) + \lambda \nabla_x \log p(x \mid y).
\end{multline}

 While this equation is exactly correct for \(\lambda=1\), it was shown in \cite{ho2022classifierfree} that using values of \(\lambda>1\) can actually yield much higher quality results. Increasing the value of lambda can be interpreted as trading off dynamics consistency with satisfying the conditioning variable. In our case, the conditioning variable \(y\), similar to \cite{ajay2023conditional}, is a discounted sum of future rewards scaled between \([0,1]\). At test time, we set the conditioning variable to 1 to guide our samples towards maximum return trajectories. In practice, we can learn both the conditional and unconditional distributions in the same model by simply masking the conditioning variable with some fixed probability. Thus at test time, performing inference with the conditioning set to the masking value will corresponding to an unconditional sample.

\subsection{Offline adaptation after dataset collection}\label{sec:offline_adapt}

When a dataset is properly labelled with input commands, we can easily recover this behaviour by adding these commands to our conditioning input. However this is often not the case, for example, when using motion capture data. Given this limitation, it is not straightforward how goal-conditioned behaviour can be recovered at test time. One approach is to treat this as an Offline RL problem, labeling the dataset with a velocity tracking reward and guiding our model towards generating samples with high returns.

We first demonstrate this is possible by defining the following reward function:

\begin{equation}\label{eq:velocity_target}
    r = e^{-3(v - v_{target})^2} - 1,
\end{equation}

where \(v_{target}\) is some fixed velocity command. We then take a discounted sum of these rewards over the next 50 timesteps to get the return. Finally, the return is transformed using equation \ref{eq:return_scaling} to make it positive 

\begin{equation}\label{eq:return_scaling}
    R = e^{R_0/A},
\end{equation}

where \(R_0\) is the un-scaled return distribution. We then linearly scale this between \([0,1]\) based on the batch statistics. The temperature of the exponential \(A\) impacts performance and is discussed further in section \ref{sec:return_scaling}.

\subsection{Diffusion for locomotion}

The overview of our model is shown in Fig \ref{fig:overview}. We take a history of \(T_{cond}\) observations, containing the robot's orientation, twist, joint position, and joint velocity. We also take the the diffusion timestep, one-hot skill vector, and return. These are all embedded with separate linear layers and concatenated to form the conditioning input. The main block of the model is a multi-headed transformer decoder. This first takes as input an embedded noise trajectory and applies causally-masked self-attention. We then apply cross-attention between this embedding and the conditioning input. This is then decoded and forms a partially denoised trajectory of future actions of length T. This process is repeated N times, eventually producing \(F_{\theta}\) from Eq.~\ref{eq:scaling} from which the fully denoised action trajectory can be computed. The first action is then applied before re-planning in a receding horizon control loop. We chose to predict multiple actions as prior work \cite{janner2022planning} has shown this helps to enforce temporally consistent trajectories. Additionally we found that using history of states was crucial for accurate system identification when deployed on hardware.


\section{Experimental results}
\label{sec:result}

We first explore the application of classifier-free guidance to offline RL as demonstrated in \cite{ajay2023conditional}. We then demonstrate the skill switching abilities of our model in conjugation with reward guidance. This is deployed onto a real robot to verify the validity of our method. We lastly ablate the effect of important hyperparameters in model, namely the guidance strength and return scaling.

\begin{figure*}[t]
    \centering
    \includegraphics[width=\linewidth]{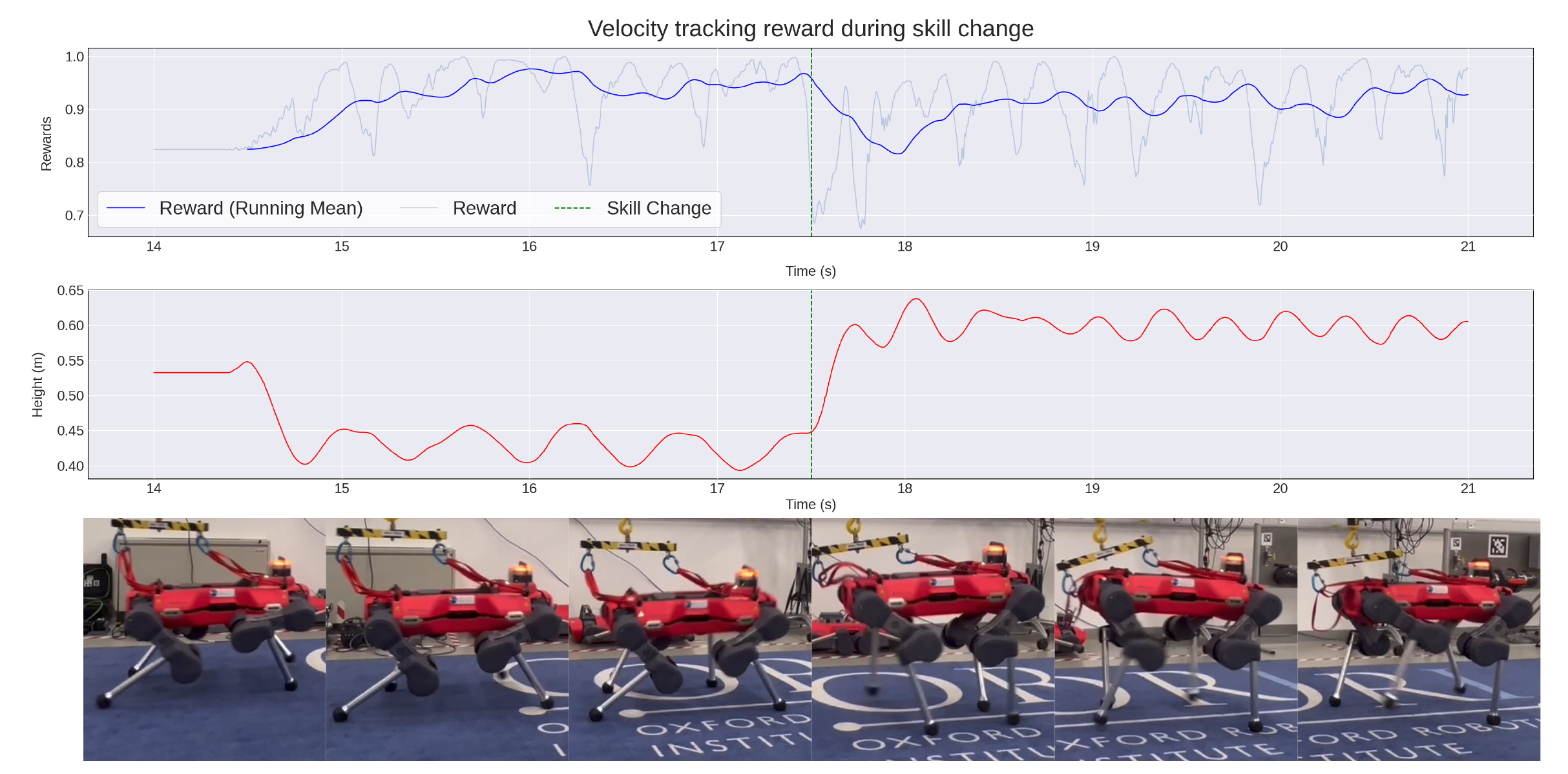}
    \caption{Velocity tracking reward, height, and snapshots from a model trained with the forward velocity reward as it switches between walking and crawling skills. The reward function is enforced by applying classifier-free guidance and robustly holds up even during the transition.}
    \label{fig:skill_switch}
    \vspace{-0.5cm}
\end{figure*}

\subsection{Dataset collection}\label{sec:dataset_collection}

To collect our dataset, we use a reinforcement learning policy trained using the reward function from \cite{gangapurwala2023learning} with random velocity commands and physical parameters. We train two separate 25Hz policies for two separate skills, walking and crawling. The training setup is almost identical for both except with different desired base heights and nominal joint positions. Both policies were trained with an action delay of a single timestep to account for the inference time of the model on hardware. We collect a total of 1M episode steps for each skill.

\subsection{Offline adaptation}\label{sec:offline_adapt_results}

After training, the model's outputs are adjusted to recover different locomotion behaviours as per Sec.~\ref{sec:offline_adapt}. Table \ref{tab:model_comp} shows the velocity tracking error of policies trained with different values of \(v_{target}\). These were chosen as they are the maximum velocities for each direction in the reference dataset: \(0.8, -0.8, 0.5\) and \(1.0 \, ms^{-1}\) respectively with all other velocities set to zero. We compare an expert model with access to the ground truth commands to our model that has no access to these commands but instead aims to maximise the reward function in equation \ref{eq:velocity_target} via classifier-free guidance. We use three different popular samplers from the SDE diffusion framework, DDPM, DDIM, and DPM++(2M) that all performed best at different tasks in prior work \cite{reuss2023goal}. 

Our results demonstrate that our method, without access to ground truth commands, can produce comparable velocity tracking to a model that does by using reward guidance. The tight error bounds on these results indicate the stability of the behaviour produced. The DDPM and DDIM samplers performed equally well, with each one slightly beating the other of different tasks. The DPM++ (2M) sampler was however significantly worse, performing much worse than the others on every task. These experiments demonstrate the effectiveness of CFG for guiding quadruped behaviour with reward functions specified after dataset collection.

\subsection{Reward tracking with skill switching}

We next investigate the skill switching capability of our model. Previous works have have looked at switching between skills with similar state distributions on smaller platforms but we wanted to test if the interpolating abilities of diffusion models would scale up to larger platforms with more dynamic transitions. To do this we collected separate datasets generated by walking and crawling reinforcement learning policies with no transitions present between the two. Our model was able to learn interpolations between these two skills which were remarkably stable over the full range of velocity commands in the dataset. The bottom row of Figure \ref{fig:skill_switch} shows snapshots from one of these transitions when deployed on real hardware.

A key benefit of diffusion-based approaches to offline RL is that since having multiple skills and applying reward guidance are independent design decisions they can both be used in conjunction. The top two rows of Figure \ref{fig:skill_switch} show the velocity tracking reward of a model trained with the forward velocity reward function and the robot's height during a skill switch. The tracking performance experiences no disruption during this switch, highlighting both the efficacy of our approaches ability to generalise into unseen transitions even under the additional constraint of being guided towards maximising some reward, and the robustness of the actions produced by guided generation.

An additional technical detail is that in order to deploy our model onto the CPU for real-time inference, we used multi-threading to avoid blocking the main processes of the robot's control loop. This allowed us to circumvent the need for additional onboard compute (e.g., a GPU) and can allow for deployment on a wider range of platforms with limited compute resources.

\subsection{Effect of lambda}

\begin{figure}
    \centering
    \includegraphics[width=\linewidth]{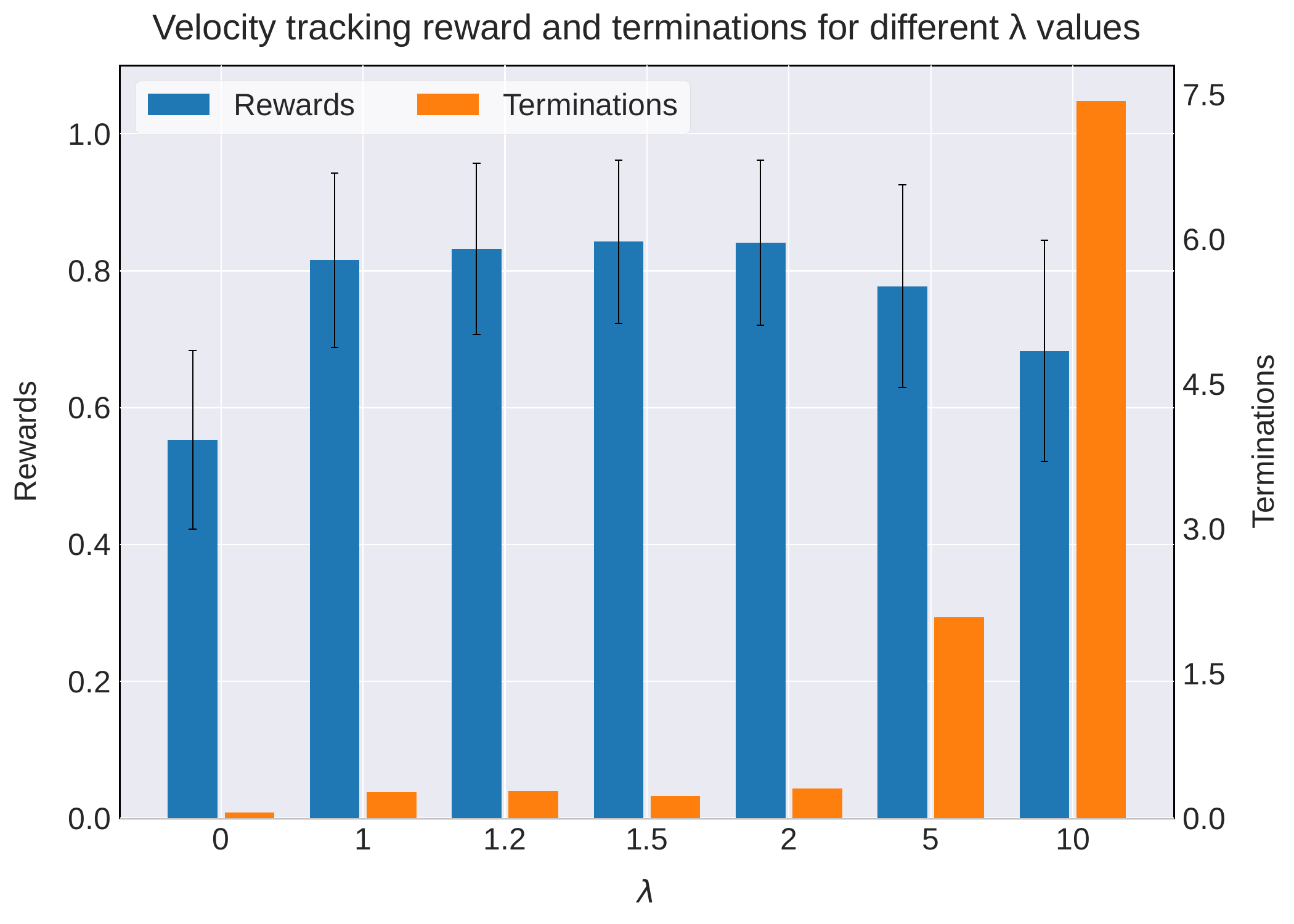}
    \caption{Velocity tracking reward and number of terminations for different guidance strength values for the forward velocity reward. Increasing guidance strength increases the performance of the mode, with the maximum performance at $\lambda=1.5$. Values much larger than 2 cause instability due to excessively trading off dynamics consistency.}
    \label{fig:lambda_comp}
    \vspace{-0.5cm}
\end{figure}

We next aim to ablate the effect of several important hyperparameters to the capabilities shown above. The first of these is the value of lambda from equation \ref{eq:cfg}. The two terms in this equations can be interpreted as generating samples that are likely given the bulk dynamics of the dataset, \(p(x)\), and satisfying the conditioning by generating samples under the conditional distribution \(p(x \mid y)\). Increasing lambda trades-off the former for the latter. Thus, we would predict that higher values of lambda generate higher reward trajectories, up to some limit, past which too much dynamics consistency is sacrificed and motion becomes unstable. The results in Figure \ref{fig:lambda_comp} shows exactly this. $\lambda = 0$ represents a sample from the unconditional model, $\lambda = 1$ from a model just using regular conditioning. Our model achieves the best results at values of 1.5 - 2 for all reward functions, showing that a small amount of reward guidance delivers better performance than just strictly conditioning. Lastly, it is also clear that for values above 2, the number of terminations quickly increases, also as per our predictions above. An interesting detail to note is that during training the $\lambda$ which gives maximum performance starts at around 10, and then as the model improves this gradually decreases to the 1.5 - 2, demonstrating that as training progresses, less guidance is needed to produce the same performance.

\subsection{Return distribution}\label{sec:return_scaling}

Another area of interest is to analyse how the shape of the return distribution affects performance. This section attempts to answer two questions. The first is whether the model just samples from a subset with returns = 1, more akin to goal conditional imitation as opposed to RL. To test this, we train a model, using the forward reward function, by just passing the returns through \ref{eq:return_scaling} and skipping the extra normalisation step afterwards to scale them between $[0,1]$, giving the returns a range of $[0.07, 0.68]$. At inference time we still condition the model on returns = 1. The first two columns of Table \ref{tab:returns} compare the average reward of an unconditional sample $(\lambda = 0)$, the maximum reward, and number of terminations of this un-scaled model to our best model from Section \ref{sec:offline_adapt}.

A few things are notable. Firstly, even when conditioned on a target return outside of its training distribution the guidance still generalises well and produces the desired effect as evidences by the increase in reward compared to the unconditional sample. Secondly, the performance is actually remarkably close to the baseline policy. However, while performance is comparable, the un-scaled policy has a much higher number of terminations due to the large size of the guidance steps from the larger deltas in the desired reward. These results, in summary highlight both the impressive generalisation of our CFG model but also the need for proper scaling of return distributions to prevent instability.

The second property to investigate is the effect of the return distribution temperature $A$. We predicted that having too narrow of a return distribution would negatively impact performance as the mode would struggle to distinguish the returns of different actions. To test this we compare the baseline policy which uses as temperature of $A=10$ to two other scaling values $A=1$ and $A=100$, listed in the last two columns of Table \ref{tab:returns}. We found that this hypothesis was indeed the case. $A=10$ empirically produces the widest return distribution (before scaling) and as such produced the best results. The other two models both performed significant worse, highlighting the important of scaling the return distribution to be as broad as possible.

\renewcommand{\arraystretch}{1.2}
\newcolumntype{b}{X}
\newcolumntype{s}{>{\hsize=.4\hsize}X}
\newcolumntype{t}{>{\hsize=.6\hsize}X}
\begin{table}
    \centering
    \begin{tabularx}{\linewidth}{bstss}
        \toprule
         Model & Baseline & No scaling & A = 1 & A = 100 \\
         \midrule 
         Unconditional Avg. Reward & 0.52 & 0.56 & 0.63 & 0.54\\
         Max Avg. Reward & 0.85 & 0.83 & 0.76 & 0.80\\
         Terminations & 0.16 & 0.57  & 0.2 & 0.53\\
         \bottomrule
    \end{tabularx}
    \caption{Average rewards and number of terminations of different models trained with different scaling return distributions. Baseline model has returns normalised between $[0,1]$ and a return temperature of $A=10$. No scaling model omits the first step and last two change the return temperature. Results are averaged over 100 environments with 250 steps each. Results are taken at the value of lambda that gave the best returns}
    \label{tab:returns}
    \vspace{-0.5cm}
\end{table}
\renewcommand{\arraystretch}{1}

\section{Conclusion}

We present a novel diffusion based method to learn a range of locomotion skills and show that we can adapt the test-time behaviour to maximise a reward after training.
This is achieved using classifier-free guidance and we demonstrate robust and high quality trajectories on the real ANYmal quadruped.
We also provide in-depth analysis of our methods limitations using ablations of several key hyperparameters.
As with all imitation learning methods, high-quality data over a broad range of environments are required for training and this could be perceived as a limitation.
However, with our classifier-free guidance, our methods are able to stitch together trajectories which maximise a desired reward after the diffusion model is trained.
Hence, we do not require labelled data, and since our method can be adapted at test time, a mixture of feasible trajectories are sufficient. 
As with the majority of methods, there is a limitation to the degree to which the model can be adjusted after training. 
Our model approximates a broad range of diverse skills and can interpolate between them, but cannot extrapolate and maximise out-of-distribution rewards.
Interesting directions for future work include using this mechanism to guarantee that the test-time behaviour remains within the training distribution, and to expand capabilities with more diverse and dynamic skills using motion capture data.







\bibliographystyle{IEEEtran}
\bibliography{IEEEabrv,references}

\end{document}

%% file: macros.tex
\newcommand{\xx}{\boldsymbol{x}}
\newcommand{\yy}{\boldsymbol{y}}
\newcommand{\zz}{\boldsymbol{z}}
\newcommand{\nn}{\boldsymbol{n}}
\newcommand{\bs}{\boldsymbol{s}}
\newcommand{\nablaxx}{\nabla_{\hspace{-0.5mm}\xx}}
\newcommand{\diff}{\mathrm{d}}